\documentclass[11pt]{article}
\usepackage{paclic34}
\usepackage{times}
\usepackage{latexsym}
\usepackage{amsmath}
\usepackage{multirow}
\usepackage{url}

\usepackage{float}
\usepackage{tikz}
\usepackage{tikz-dependency}
\usepackage{pgfplots}
    \pgfplotsset{compat=1.8}
    \usepgfplotslibrary{statistics}
\usepackage{subcaption}
\usepackage{soul}
\usepackage{xcolor}
    \definecolor{lightgray}{rgb}{0.827, 0.827, 0.827}
    \sethlcolor{lightgray}
\usepackage[a4paper]{geometry}
\title{Zero-shot and few-shot approaches for tokenization, tagging, and dependency parsing of Tagalog text}

\author{Angelina Aquino {\normalfont and} Franz de Leon \\
    Digital Signal Processing Laboratory, Electrical and Electronics Engineering Institute \\
    University of the Philippines, Diliman, Quezon City, Philippines \\
    \texttt{\{angelina.aquino, franz.de.leon\}@eee.upd.edu.ph} \\
}

\date{}

\begin{document}
\maketitle
\begin{abstract}
    The grammatical analysis of texts in any written language typically involves a number of basic processing tasks, such as tokenization, morphological tagging, and dependency parsing. State-of-the-art systems can achieve high accuracy on these tasks for languages with large datasets, but yield poor results for languages which have little to no annotated data. To address this issue for the Tagalog language, we investigate the use of alternative language resources for creating task-specific models in the absence of dependency-annotated Tagalog data. We also explore the use of word embeddings and data augmentation to improve performance when only a small amount of annotated Tagalog data is available. We show that these zero-shot and few-shot approaches yield substantial improvements on grammatical analysis of both in-domain and out-of-domain Tagalog text compared to state-of-the-art supervised baselines.
\end{abstract}

\section{Introduction}\label{sec-intro}

\begin{figure*}[ht]
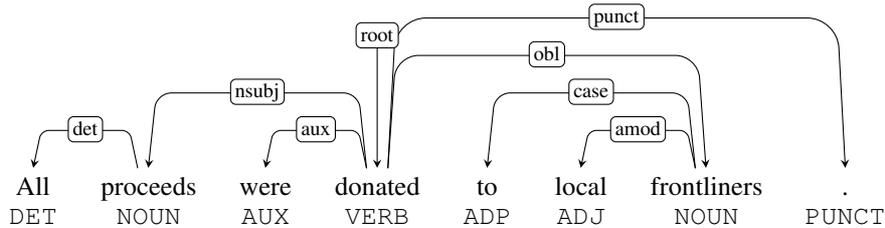

        \centering
        \begin{dependency}
            \begin{deptext}[font=\small, column sep=0.4cm]
                All \&
                proceeds \&
                were \&
                donated \&
                to \&
                local \&
                frontliners \&
                .\\
                \texttt{DET} \&
                \texttt{NOUN} \&
                \texttt{AUX} \&
                \texttt{VERB} \&
                \texttt{ADP} \&
                \texttt{ADJ} \&
                \texttt{NOUN} \&
                \texttt{PUNCT}\\
            \end{deptext}
            \depedge{2}{1}{det}
            \depedge{4}{2}{nsubj}
            \depedge{4}{3}{aux}
            \depedge{7}{5}{case}
            \depedge{7}{6}{amod}
            \depedge{4}{7}{obl}
            \depedge{4}{8}{punct}
            \deproot{4}{root}
        \end{dependency}
        \caption{Grammatical analysis (using the Universal Dependencies annotation format\footnotemark) for the sentence: \textit{All proceeds were donated to local frontliners.}}
        \label{fig:deptree}
    \end{figure*}

    The grammar of a language is the set of rules and processes which govern the composition of words and sentences in that language. When conducting a grammatical analysis of sentences in a given text, we typically need to perform several tasks such as tokenization, part-of-speech tagging, and syntactic parsing, each of which pertains to a distinct level of grammatical structure. These three levels of analysis are illustrated in Fig. \ref{fig:deptree} as a dependency tree, one of the most common means of representing a sentence's grammatical structure.

    With the advent of machine learning, many natural language processing (NLP) pipelines can now accomplish such analysis tasks with high accuracy for languages with large annotated datasets. However, these systems still yield poor results in a low-resource setting, for languages which have little to no available dependency-annotated data \cite{zeman-etal-2018-conll}. This renders the outputs of such pipelines unsuitable for use in downstream tasks and limits the development of more advanced technologies for these languages.

    This is not to say that a language which lacks annotated data is inherently a ``low-resource language''---on the contrary, many such languages do in fact possess a wide range of other language resources, including literary works (e.g. books, newspapers, poetry), linguistic references (e.g. dictionaries, grammar textbooks), and the expert knowledge of those proficient with the language. Unlike annotated datasets, these resources cannot be directly fed into existing NLP pipelines as inputs, but it is still possible to encode the information from these resources into usable formats. Given this scenario, we thus ask:
    \begin{enumerate}
        \item Can we create a pipeline for grammatical analysis \textit{without} any dependency-annotated data, using only these alternative language resources?
        \item Can we improve the performance of existing pipelines by using alternative resources and methods \textit{together with} a small amount of dependency-annotated data?
    \end{enumerate}

    In this work, we explore various approaches to achieve these two objectives for the Tagalog language (Section \ref{sec-methods}), evaluating their performance on educational and news text against state-of-the-art baselines (Section \ref{sec-eval}). Using the best-performing approaches, we develop a few-shot pipeline and a zero-shot pipeline (Section \ref{sec-pipelines}) which both improve on the current state-of-the-art, and which serve as competent benchmarks for the automated grammatical analysis of Tagalog text. We conduct this work using the Universal Dependencies (UD) framework, which provides cross-linguistically consistent guidelines for grammatical annotation in many languages \cite{nivre-etal-2016-universal}. We situate our work against prior endeavors in automated grammatical analysis for Tagalog, and suggest directions for future research (Section \ref{sec-discussion}). Our scripts, models, and datasets described in this paper are freely available for general use.\footnotemark

    \footnotetext{\url{https://universaldependencies.org/guidelines.html}}
    \footnotetext{\url{https://github.com/AngelAquino/tagalog-ud-pipeline}}

\section{Candidate approaches}\label{sec-methods}

    Existing NLP pipelines typically perform several annotation tasks through a chain of individual processing components, each of which can be improved in a low-resource setting using alternative methods and resources. We tested several approaches for implementing the individual pipeline components, then selected the best-performing approaches for each task to create our new pipelines. A description of the specific annotation tasks, and the candidate approaches we have chosen to test for each task, are described in the sections below.

    \textbf{Tokenization} is the task of segmenting running text into distinct levels of organization, such as tokens, words, and sentences. We implemented and evaluated the following alternative approach for tokenization:
        \begin{itemize}
            \item \textit{Unsupervised tokenization.} Since tokenization is one of the first preprocessing tasks performed for any text analysis, many tokenizers are included in widely-used NLP libraries. For this task, we used the Python NLTK \cite{bird-2006-nltk} implementation of the Punkt tokenizer \cite{kiss-strunk-2006-unsupervised}, which uses heuristic boundary detection algorithms fine-tuned on the American English \textit{Wall Street Journal} corpus.
        \end{itemize}

    \textbf{Part-of-speech tagging} is the task of identifying the parts of speech (e.g. noun, verb, adjective) of words or tokens in a sentence. We implemented and evaluated the following approaches for part-of-speech tagging:
        \begin{itemize}
            \item \textit{Tag conversion.} A POS-tagged corpus for Tagalog was previously developed by Nocon and Borra \shortcite{nocon-borra-2016-smtpost} using the language-specific MGNN tagset. We developed scripts to map this tagset to the UD POS tagset, and used the corpus text together with the converted tags to train a POS tagger model.
            \item \textit{Multilingual annotation projection.} In the absence of labeled data for a target language, annotations such as POS tags can be projected from a high-resource source language to a low-resource target language through parallel text. Several massively multi-parallel corpora with Tagalog coverage now exist \cite{agic-etal-2016-multilingual,agic-vulic-2019-jw300}. We adopted the methodology by
            Plank and Agi\'{c} \shortcite{plank-agic-2018-distant} for part-of-speech projection.
        \end{itemize}

    \textbf{Morphological analysis}. The UD framework includes levels of annotation for the lemmas (i.e. canonical or uninflected forms) and morphological features (e.g. gender, number, tense) of each word. We implemented and evaluated the following approach for annotating these properties:
        \begin{itemize}
            \item \textit{Rule-based analysis.} Several rule-based methods for morphological analysis of Tagalog were previously implemented \cite{roxas-mula-2008-morphological,cheng-etal-2017-mag-tagalog} which have primarily focused on the language's complex verbal morphology. We formed a similar set of rules to identify lemmas and selected UD features, and implemented them as a finite-state transducer.
        \end{itemize}

    \textbf{Dependency parsing} is the task of determining the correct head-dependent relationships between words in a given sentence, as well as the labels for each of these relations. We implemented and evaluated the following approaches for dependency parsing:
        \begin{itemize}
            \item \textit{Multilingual annotation projection.} Similar to what was described for POS tagging, annotation projection can also be applied to dependency parsing. Here we implemented the methodology of Agi\'{c} et al. \shortcite{agic-etal-2016-multilingual} for projection of dependency trees.
            \item \textit{Data augmentation.} When only a small amount of labeled dependency data is available for a given language, several operations can be performed to artificially augment the number of available sentences at training time. In line with the work of Vania et al. \shortcite{vania-etal-2019-systematic}, we implemented sentence morphing (i.e. cropping and rotating existing sentences) to produce additional training data.
        \end{itemize}

    \textbf{Neural system with embeddings}. The Stanza neural pipeline developed by the Stanford NLP Group has produced state-of-the-art results for dependency parsing for many languages, and for low-resource settings in particular \cite{qi-etal-2018-universal}. This pipeline consists of a series of neural architectures, and makes use of pre-trained word embeddings (i.e. high-dimensional vector representations encoding the context of words in a given language, trained using large amounts of raw text). We trained Stanza models for tokenization, part-of-speech tagging, morphological analysis, and dependency parsing using the UD-annotated Ugnayan treebank \cite{aquino-de-leon-2020-parsing} and the fastText word embeddings for Tagalog \cite{bojanowski-etal-2017-enriching}, and compared the performance of these models to those of the other candidate approaches for the pipeline.

    Each of these implementations was evaluated alongside the baseline performance of the UDPipe monolingual Tagalog model, which was determined to be the best supervised system for Tagalog UD parsing among those we evaluated in a previous study \shortcite{aquino-de-leon-2020-parsing}. We also used UDify (a state-of-the-art language-agnostic parsing pipeline) as a zero-shot baseline, as a system which achieved competitive performance on Tagalog tagging and parsing without the use of any Tagalog UD annotations in its training.

    We tested all approaches using an expanded version of the Ugnayan treebank for Tagalog. The version of this treebank that is currently available on the Universal Dependencies repository\footnote{\url{https://universaldependencies.org/}} consists of 94 sentences (1097 words) of educational text from fiction and nonfiction resources of the DepEd Learning Resource portal. The treebank has been further expanded to include an additional 25 sentences (792 words) of news text from articles by the Philippine Information Agency. We used the educational section of the treebank as training data where required, and the news section purely as testing data.

\section{Evaluations}\label{sec-eval}

    \begin{table*}[ht]
        \small
        \centering
        \begin{tabular}{ll|ccc|ccc}
            \hline
            & & \multicolumn{3}{c|}{\textit{a. Educational text (raw)}} & \multicolumn{3}{c}{\textit{b. News text (raw)}} \\
            \textsc{Method} & \textsc{Token} & \textsc{Word} &\textsc{Sent} & \textsc{Token} &\textsc{Word} &\textsc{Sent} \\
            \hline
            \textit{few-shot} & UDPipe (supervised baseline) & \textbf{99.27} & \textbf{95.67} & \textbf{95.41} & 91.03 & 87.81 & 66.67 \\
            & Stanza (supervised + embeds) & 99.11 & 93.74 & 93.36 & 93.59 & 86.97 & 50.91 \\
            \hline
            \textit{zero-shot} & Punkt (unsupervised) & 97.25 & 85.08 & 89.69 & \textbf{97.10} & \textbf{87.58} & \textbf{76.92} \\
            \hline
        \end{tabular}
        \caption{F$_1$ scores on tokenization from raw text, tested on both the educational and news sections of the Ugnayan treebank. Results in (a) for both UDPipe and Stanza were averaged over 10-fold cross-validation. \textbf{Bold}: highest scores for each test set.}
        \label{tab:pipeline-tok}
    \end{table*}

    \textbf{Tokenization.} For the task of tokenization, we compare the performance of the UDPipe baseline tokenizer to two alternative systems:
    \begin{enumerate}
        \item The tokenizer component of a Stanza pipeline trained on Ugnayan data as well as fastText Tagalog word embeddings, and
        \item the unsupervised Punkt tokenizer, as implemented in Python NLTK.
    \end{enumerate}

    We report the F$_1$ scores for each of these models in Table \ref{tab:pipeline-tok}. We find that the supervised UDPipe and Stanza tokenizers both outperform the Punkt tokenizer on word and sentence tokenization when tested on in-domain (educational) text, with the UDPipe tokenizer giving marginally better results. On the other hand, the Punkt tokenizer surpasses both UDPipe and Stanza on sentence tokenization of out-of-domain (news) text by a large margin.

    Upon inspection of the predicted tokenizations for each system, we find that the Punkt tokenizer gives poor sentence segmentation for quotations (e.g. \textit{"Maraming salamat po!" sabi ko.}) whereas the UDPipe and Stanza tokenizers tend to incorrectly split sentences based on periods following abbreviations (e.g. \textit{Engr.}). These may be attributed to the nature of the different models: the UDPipe and Stanza models were trained on the educational portion of Ugnayan, which does not contain any occurrences of abbreviations, so the model tends to recognize all periods as sentence boundaries, regardless of context. In contrast, the Punkt tokenizer distinguishes periods that follow abbreviations from sentence-terminating periods with high accuracy, but may fail to disambiguate other sentence-terminating punctuation marks which are instead located in the middle of a sentence (as with quotations). These properties may explain the differences in performance between the models on the two datasets: the educational dataset contains $7$ sentences with quotations and none with abbreviations (out of $94$), while the news dataset contains $4$ sentences with abbreviations containing periods (out of $25$).

    \begin{table*}[ht]
        \small
        \centering
        \begin{tabular}{ll|ccc|ccc}
            \hline
            & & \multicolumn{3}{c|}{\textit{a. Educational text}} & \multicolumn{3}{c}{\textit{b. News text}} \\
            & & \multicolumn{3}{c|}{\textit{(tokenized)}} & \multicolumn{3}{c}{\textit{(tokenized)}} \\
            & \textsc{Method} & \textsc{UPOS} & \textsc{Feat} &\textsc{Lemm} & \textsc{UPOS} & \textsc{Feat} &\textsc{Lemm} \\
            \hline
            \textit{few-shot} & UDPipe (supervised baseline) & 83.76 & 94.21 & 89.79 & 68.21 & \textbf{94.19} & 75.62 \\
            & Stanza (supervised + embeds) & \hl{\textbf{91.16}} & \hl{\textbf{95.53}} & \hl{\textbf{92.68}} & \hl{\textbf{74.38}} & 94.07 & \hl{\textbf{79.02}} \\
            \hline
            \textit{zero-shot} & UDify (zero-shot baseline) & 59.80 & 65.45 & 71.01 & \textbf{61.11} & 73.48 & \textbf{72.47} \\
            & POS tag conversion (MGNN) & \textbf{68.19} & & & 57.07 & & \\
            & POS projection (en) & 61.17 & & & 52.53 & & \\
            & POS projection (en+id+it+pl) & 61.90 & & & 57.20 & & \\
            & Foma FST (v1) & & 93.53 & 71.01 & & \hl{\textbf{95.33}} & 70.96 \\
            & Foma FST (v2) & & \textbf{93.71} & \textbf{71.19} & & \hl{\textbf{95.33}} & 70.96 \\
            \hline
        \end{tabular}
        \caption{F$_1$ scores on part-of-speech tagging, morphological feature analysis, and lemmatization from tokenized text, tested on both the educational and news sections of the Ugnayan treebank. Results in (a) for both UDPipe and Stanza were averaged over 10-fold cross-validation. \textbf{Bold}: highest scores per system type. \hl{Gray}: highest scores across all models.}
        \label{tab:pipeline-tag-morph}
    \end{table*}

    \textbf{Tagging and morphology.} For part-of-speech tagging, we compare the performance of the baseline taggers to four alternatives:
    \begin{enumerate}
        \item The POS tagger component of a Stanza pipeline trained on Ugnayan data as well as fastText Tagalog word embeddings,
        \item A POS tagger trained using the ISIP-SAFE Part-of-Speech Tagger corpus, with tags converted from the MGNN tagset into the UD tagset,
        \item A POS tagger trained on tags projected from a single source language (English) to Tagalog via alignment of a random selection of parallel English-Tagalog sentences, and
        \item A POS tagger trained on tags projected from four source languages (English, Italian, Polish, and Indonesian) to Tagalog via alignment of parallel sentences with optimal coverage across all source-target language pairs.
    \end{enumerate}

    We used the Bilty bidirectional long-short term memory tagger \cite{plank-etal-2016-multilingual} to train models 2 to 4. We implemented the tag conversion from the MGNN tagset to the UD tagset as a Python script. We performed this conversion on the ISIP-SAFE Part-of-Speech corpus, and used the corpus text and converted tags to train model 2.

    All parallel sentences used in projection for models 3 and 4 come from the JW300 corpus \cite{agic-vulic-2019-jw300}. For model 3, we selected English for single-source projection as the source language with the highest number of parallel sentences to Tagalog in JW300; we used a randomized selection of text covering approximately 100,000 words (contained in 4831 sentences) to train the model. For model 4, we selected four source languages to represent four different language groups --- English for Germanic, Italian for Romance, Polish for Slavic, and Indonesian for Austronesian (to which Tagalog also belongs); here we used the top 5000 Tagalog sentences with optimal alignment coverage averaged across the four source languages.

    For models 3 and 4, we trained the source-side taggers using the English-GUM, Italian-ITSD, Polish-PDB, and Indonesian-GSD treebanks from UD v2.6, and used these to tag the source sentences. We then aligned the source and target words using the Eflomal alignment tool \cite{ostling-tiedemann-2016-efficient}, determined the optimal tag for each target word as the tag with the highest sum of tagger confidences across all aligned source words, then projected these optimal tags onto the target words. The target Tagalog words and projected tags were then used as training data for these models.

    For morphological feature annotation and lemmatization, we compare the performance of the baselines to three alternatives:
    \begin{enumerate}
        \item The lemmatizer component of a Stanza pipeline trained on Ugnayan data as well as fastText Tagalog word embeddings,
        \item Version 1 of our Foma finite-state transducer for Tagalog morphology, which specifies rules for verbal reduplication and the occurrence of verbal affixes \textit{-um-} and \textit{-in-}, and
        \item Version 2 of the Foma FST for Tagalog, which includes all of the rules from Version 1, and adds rules for the occurrence of the verbal affixes \textit{-nag-} and \textit{-hin-}.
    \end{enumerate}

    We report the F$_1$ scores for each of these models in Table \ref{tab:pipeline-tag-morph}. We find that for part-of-speech tagging and lemmatization, Stanza outperforms both baseline systems on both educational and news texts, while both Stanza and the baseline UDPipe system outperform each of the other alternatives on both datasets by a wide margin. For morphological feature annotation, the Stanza and UDPipe systems also yield marginally better feature annotation than the Foma FST models on the in-domain educational text, but the FST models give a slight improvement over their performance for the out-of-domain news text; all three approaches surpass the zero-shot UDify system on this task by a wide margin. In the absence of UD-annotated Tagalog data, both POS tag conversion and POS projection yield comparable performance to the UDify system, with the tag conversion model and the UDify model performing slightly better on the educational text and news text, respectively.

    We can observe that for the part-of-speech tagging task, the use of multiple source languages in the projection approach seems to improve performance over the use of only a single source language; this improvement may be attributed to the coverage-based selection of sentences in the multiple-source model (as opposed to random selection for the single-source model), since a greater amount of coverage has been shown to improve part-of-speech projection performance in prior works \cite{duong-etal-2013-simpler,plank-agic-2018-distant}. Further study may be needed to determine the general effects of the number of source languages, as well as their similarity to the target language, on the accuracy of the projection method.

    We also note both Versions 1 and 2 of our Tagalog FST yielded the same performance on the news dataset. This indicates that the additional rules included in Version 2 did not correspond to any verbal phenomena that were present in the news dataset, and that in general, the addition of rules may or may not result in an improvement in the morphological analysis of a given text, depending on the contents of that text. Nevertheless, we have shown that even a small number of rules (23 for Version 1, and 27 for Version 2) can account for a large number of verbal phenomena in Tagalog corpora. Moreover, the behavior of such a rule-based system is highly explainable and can be easily expanded to account for any well-defined morphological process (as opposed to the neural models used in the other approaches, which have low explainability, and which can be trained on several examples of some morphological phenomenon without any guarantees that the model is able to generalize to other occurrences of the same phenomenon).

    \textbf{Dependency parsing.} For the dependency parsing task, we compare the performance of the baselines to four alternatives:
    \begin{enumerate}
        \item The dependency parser component of a Stanza pipeline trained on Ugnayan data as well as fastText Tagalog word embeddings,
        \item The dependency parser component of a UDPipe model trained on Ugnayan data augmented by sentence morphing,
        \item The dependency parser component of a UDPipe model trained on dependency trees projected from a single source language (English) to Tagalog via alignment of parallel sentences with optimal coverage, and
        \item The dependency parser component of a UDPipe model trained on dependency trees projected from four source languages (English, Italian, Polish, and Indonesian) to Tagalog via alignment of parallel sentences with optimal coverage across all source-target language pairs.
    \end{enumerate}

    For model 2, we implemented a program which creates additional training sentences through sentence morphing. This program selects all sentences with both \texttt{nsubj} and \texttt{obj} dependents of the \texttt{root} (i.e. subject and direct object clauses of the predicate). It then identifies the word order used in each selected sentence---either \texttt{VSO}, \texttt{SVO}, or \texttt{VOS}, which are the three grammatical word orders in Tagalog---and generates two additional sentences corresponding to the other two word orders by swapping the positions of the clauses in the dependency tree. The educational portion of the Ugnayan treebank was passed as input to this program, and the morphs generated by the program were appended to the original set of sentences, together forming the training data for the parser model.

    \begin{table*}[ht]
        \small
        \centering
        \begin{tabular}{ll|cc|cc}
            \hline
            & & \multicolumn{2}{c|}{\textit{a. Educational text}} & \multicolumn{2}{c}{\textit{b. News text}} \\
            & & \multicolumn{2}{c|}{\textit{(tagged)}} & \multicolumn{2}{c}{\textit{(tagged)}} \\
            & \textsc{Method} & \textsc{UAS} & \textsc{LAS} & \textsc{UAS} & \textsc{LAS} \\
            \hline
            \textit{few-shot} & UDPipe (supervised baseline) & 75.50 & 68.95 & 53.28 & 43.06 \\
            & Stanza (supervised + embeds) & 75.08 & 66.09 & 51.01 & 36.99 \\
            & data augmentation (morph) & \hl{\textbf{77.33}} & \hl{\textbf{71.60}} & \hl{\textbf{58.33}} & \hl{\textbf{47.85}} \\
            \hline
            \textit{zero-shot} & UDify (zero-shot baseline) & \textbf{51.96} & \textbf{32.18} & \textbf{53.28} & \textbf{36.62} \\
            & projection (en) & 26.62 & 21.33 & 26.77 & 18.81 \\
            & projection (en+id+it+pl) & 28.08 & 20.69 & 24.24 & 14.77 \\
            \hline
        \end{tabular}
        \caption{F$_1$ scores on dependency parsing from gold-tagged text, tested on both the educational and news sections of the Ugnayan treebank. Results in (a) for the UDPipe baseline, Stanza, and UDPipe data augmentation models were averaged over 10-fold cross-validation. \textbf{Bold}: highest scores per system type. \hl{Gray}: highest scores across all models.}
        \label{tab:pipeline-parse}
    \end{table*}

    For models 3 and 4, we used the same set of 5000 Tagalog sentences (and the source sentences parallel to them) as in the multiple-source part-of-speech projection described earlier. We trained source-side parsers using the English-GUM, Italian-ITSD, Polish-PDB, and Indonesian-GSD treebanks from UD v2.6, and used these to generate dependency trees for the source sentences. We then modified the Eflomal alignment tool to print alignment probabilities per word (since the tool originally provided only sentence alignment probabilities in its output), and used this to generate word alignments and probabilities per source-target sentence pair. The parallel sentences and their corresponding alignments were used as inputs to the projection tools developed by Agi\'{c} et al. \shortcite{agic-etal-2016-multilingual} to generate dependency trees projected onto the target sentences through probability weighting and directed maximum spanning tree decoding.

    Since these tools only projected dependency edges but not dependency labels, we needed to develop an additional method to predict dependency labels for the projected edges using only available data from the source languages. For this, we trained delexicalized random forest classifiers which predict the dependency label for a word using two features: (1) the POS tag of that word, and (2) the POS tag of the word which heads it. (We say that these models are delexicalized since they only use POS tags as inputs, not the word forms themselves.) We used the same four source-side treebanks as input data for these classifiers. We then took the projected POS tags from our earlier pipeline tests, and used these together with the projected edges in order to predict dependency labels for the target sentences. Finally, we used the trees composed of projected edges, projected POS tags, and predicted dependency labels to train models 3 and 4 above.

    We report the F$_1$ scores for each of these models in Table \ref{tab:pipeline-parse}. Among the supervised systems, the UDPipe model trained on augmented data yielded the best UAS and LAS on both datasets, while the Stanza system scored the lowest. Among the zero-shot systems, the UDify model outperformed both projection models by a wide margin. There was a sizeable gap in attachment scores between the supervised and zero-shot models on the educational (in-domain) set; on the other hand, the UDify models yielded scores that approached or even exceeded those of the supervised systems on the news (out-of-domain) set.

    We observe that between the two projection models, the single-language model performed slightly better (in terms of LAS on the educational set, and both UAS \& LAS on the news set) than the multiple-language model. This runs in contrast to the part-of-speech tagging task, where we found that multiple-source POS projection gave better accuracy than the single-source alternative. Although no study to our knowledge has explored the relationship between the number of source languages and the accuracy of the resulting dependency projection, prior works have shown that the best projection approach, and the best source languages, vary across different target languages \cite{johannsen-etal-2016-joint,agic-etal-2016-multilingual}. Further work may be needed to determine the effect of each of these parameters on projection for Tagalog and for other Philippine languages.

    Beyond the results shown here, we have also observed that although the Stanza model yielded slightly lower average scores than the baseline UDPipe model when parsing from gold-tagged text, the Stanza model outperforms the UDPipe baseline on average when parsing using system-predicted part-of-speech tags and morphological annotations. We initially found this result counter-intuitive, since both parsers were configured to use system-predicted tags during their training, and we have shown in the previous section that the Stanza system produces more accurate tags (i.e. tags that are closer overall to the gold tags) than the UDPipe parser. Hence, we expected that the Stanza parser similarly outperforms the UDPipe parser on gold-tagged data. However, once we inspected the scores of both systems on each of the 10 cross-validation folds for the education set, we found that Stanza in fact gave better UAS for 5 out of 10 folds, and equal or better LAS for 4 out of 10 folds. (In comparison, the augmented model gives better or equal UAS than the UDPipe baseline for 8 out of 10 folds, and better or equal LAS for 9 out of 10 folds.) We therefore conclude that the Stanza and UDPipe baseline models give comparatively equal parsing performance, whereas data augmentation tends to improve on both of these systems overall.

    \begin{table*}
        \small
        \centering
        \begin{tabular}{l|ccc|ccc|cc}
            \hline
            \textit{a. Educational text (raw)} & \multicolumn{3}{c|}{\textit{Tokenization}} & \multicolumn{3}{c|}{\textit{Tagging}} & \multicolumn{2}{c}{\textit{Parsing}} \\
            \textsc{Pipeline} & \textsc{Token} & \textsc{Word} &\textsc{Sent} & \textsc{UPOS} & \textsc{Feat} & \textsc{Lemm} & \textsc{UAS} &\textsc{LAS} \\
            \hline
            few-shot & \textbf{98.75} & \textbf{93.56} & \textbf{98.57} & \textbf{85.12} & \textbf{90.55} & \textbf{87.21} & \textbf{67.29} & \textbf{60.15} \\
            zero-shot
            & 97.25 & 85.08 & 89.69 & 59.57 & 78.75 & 66.57 & 48.53 & 33.29 \\
            cross-lingual & 97.40 & 85.22 & 92.38 & 27.95 & 46.56 & 65.92 & 18.78 & 9.41 \\
            \hline
            \multicolumn{9}{c}{ } \\
            \hline
            \textit{b. News text (raw)} & \multicolumn{3}{c|}{\textit{Tokenization}} & \multicolumn{3}{c|}{\textit{Tagging}} & \multicolumn{2}{c}{\textit{Parsing}} \\
            \textsc{Pipeline} & \textsc{Token} & \textsc{Word} &\textsc{Sent} & \textsc{UPOS} & \textsc{Feat} & \textsc{Lemm} & \textsc{UAS} &\textsc{LAS} \\
            \hline
            few-shot & 90.85 & 84.15 & 32.14 & \textbf{61.51} & 78.45 & 65.98 & 35.26 & 26.70 \\
            zero-shot & \textbf{97.29} & \textbf{87.74} & 83.02 & 53.77 & \textbf{83.80} & 66.89 & \textbf{44.07} & \textbf{32.92} \\
            cross-lingual & 96.64 & 87.13 & \textbf{86.89} & 30.77 & 55.05 & \textbf{67.29} & 10.12 & 5.24 \\
            \hline
        \end{tabular}
        \caption{F$_1$ scores on tokenization, tagging, and parsing from raw text, tested on both the educational and news sections of the Ugnayan treebank. \textbf{Bold}: highest scores for each test set.}
        \label{tab:pipeline-final}
    \end{table*}

\section{Final pipelines}\label{sec-pipelines}

    For our final pipelines, we differentiate between two low-resource data settings: few-shot (i.e. when a small amount of UD-annotated training data is available), and zero-shot (i.e. when no UD-annotated data is available). Based on our experiments above, we propose the use of the following pipelines in each setting:
    \begin{enumerate}
        \item \textit{Few-shot.} The Stanza pipeline, trained with word embeddings and UD-annotated data augmented through sentence morphing
        \item \textit{Zero-shot.} A pipeline consisting of: the Punkt unsupervised tokenizer; the Bilty POS tagger, trained on tags converted from existing resources for the target language (e.g. a dictionary or a tagged corpus); the Foma morphological analyzer, compiled with rules written based on existing grammar texts for the target language; and the UDify universal parser model
    \end{enumerate}
    For our final evaluation, we compare the performance of these two pipelines to the UDPipe Indonesian-GSD model, which was the cross-lingual model with the best LAS among those we evaluated in a previous study \shortcite{aquino-de-leon-2020-parsing}.

    We report the F$_1$ scores for each of these models in Table \ref{tab:pipeline-final}. When testing on educational text, we find that the few-shot Stanza pipeline (trained in-domain) outperforms both the zero-shot and cross-lingual pipelines on all tokenization, tagging, and parsing tasks. The results on the news data, on the other hand, are more varied:
    \begin{itemize}
        \item For token- and word-level tokenization, the zero-shot pipeline gives the highest average scores, performing slightly better than the cross-lingual pipeline, with the few-shot pipeline lagging behind on token-level segmentation.
        \item For sentence-level tokenization, the cross-lingual pipeline gives the best performance, followed closely by the zero-shot pipeline, with both systems far surpassing the few-shot pipeline (trained out-of-domain).
        \item For part-of-speech tagging and morphological feature annotation, the gaps between scores of each of the systems are more pronounced. The few-shot pipeline does best on the POS task, the zero-shot pipeline does best on feature annotation, and the cross-lingual pipeline is well behind the other two on both tasks.
        \item For lemmatization, the cross-lingual pipeline achieves the best average, but the other two systems give roughly similar performance on the task.
        \item For dependency parsing, the zero-shot pipeline outperforms the few-shot pipeline on both unlabeled and labeled attachment, with both systems outperforming the cross-lingual pipeline by a large margin.
    \end{itemize}

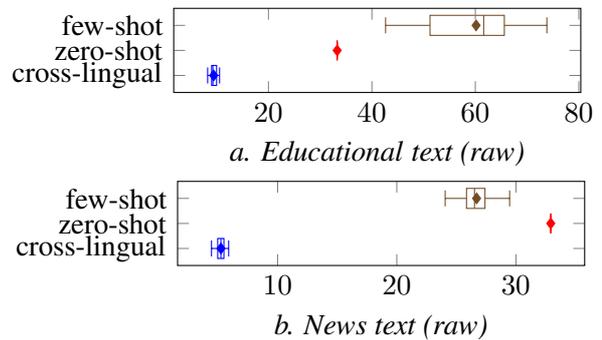
\begin{figure}[ht]
        \centering
        \begin{tikzpicture}
          \begin{axis}
            [
            ytick={1,2,3},
            yticklabels={cross-lingual,zero-shot,few-shot},
            xlabel={\textit{a. Educational text (raw)}},
            axis on top,
            width=0.9\linewidth,
            height=0.35\linewidth,
            ]
            \addplot+[
            boxplot prepared={
              lower whisker=8.25,
              lower quartile=9.02,
              median=9.26,
              upper quartile=9.98,
              upper whisker=10.56,
              average=9.41,
            },
            ] coordinates {};
            \addplot+[
            boxplot prepared={
              lower whisker=33.29,
              lower quartile=33.29,
              median=33.29,
              upper quartile=33.29,
              upper whisker=33.29,
              average=33.29,
            },
            ] coordinates {};
            \addplot+[
            boxplot prepared={
              lower whisker=42.65,
              lower quartile=51.24,
              median=61.615,
              upper quartile=65.61,
              upper whisker=73.83,
              average=60.15,
            },
            ] coordinates {};
          \end{axis}
        \end{tikzpicture}
        \begin{tikzpicture}
          \begin{axis}
            [
            ytick={1,2,3},
            yticklabels={cross-lingual,zero-shot,few-shot},
            xlabel={\textit{b. News text (raw)}},
            width=0.9\linewidth,
            height=0.35\linewidth,
            ]
            \addplot+[
            boxplot prepared={
              lower whisker=4.45,
              lower quartile=4.97,
              median=5.24,
              upper quartile=5.49,
              upper whisker=5.89,
              average=5.24,
            },
            ] coordinates {};
            \addplot+[
            boxplot prepared={
              lower whisker=32.92,
              lower quartile=32.92,
              median=32.92,
              upper quartile=32.92,
              upper whisker=32.92,
              average=32.92,
            },
            ] coordinates {};
            \addplot+[
            boxplot prepared={
              lower whisker=24.06,
              lower quartile=25.85,
              median=26.52,
              upper quartile=27.40,
              upper whisker=29.47,
              average=26.70,
            },
            ] coordinates {};
          \end{axis}
        \end{tikzpicture}
        \caption{LAS spreads for 10-fold cross-validation on tokenization, tagging, and parsing from raw text, tested on the educational section of the Ugnayan treebank.}
        \label{fig:pipeline-final-boxplot}
    \end{figure}

    Since our focus is on dependency parsing, we use the labeled attachment score as the primary measure of system performance. For this, we compare the ranges of labeled attachment scores achieved by each model across 10 folds of cross-validation, wherein the training and dev partitions are partitioned differently for each fold. (None of the zero-shot components require train/dev partitions in their development, so only a single model is used per component across all tests, and only a single score is reported for each test set.) These results are visualized in Fig. \ref{fig:pipeline-final-boxplot}.

    We find that when testing in-domain (on educational text), the few-shot pipeline yields higher scores across all folds than the zero-shot pipeline. The opposite is true when testing out-of-domain (on news text): here, the zero-shot pipeline outperforms the few-shot pipeline on all folds. We can see that the performance of the zero-shot system stays consistent across both types of data (with LAS between $32\%$ and $34\%$ for both). In contrast, the performance of the few-shot system is highly sensitive to the train-dev split used at training time (with a difference of over $30\%$ LAS between the lowest and highest scores achieved on the educational text, and to the type of data encountered at training versus testing (with the median LAS dropping by over $30\%$ from the in-domain to the out-of-domain scenario). We therefore recommend the use of the zero-shot pipeline for general Tagalog parsing purposes, and would only consider using the few-shot pipeline if the data to be parsed is in the same domain as any annotated datasets available for training the pipeline.

    \begin{figure*}[ht]
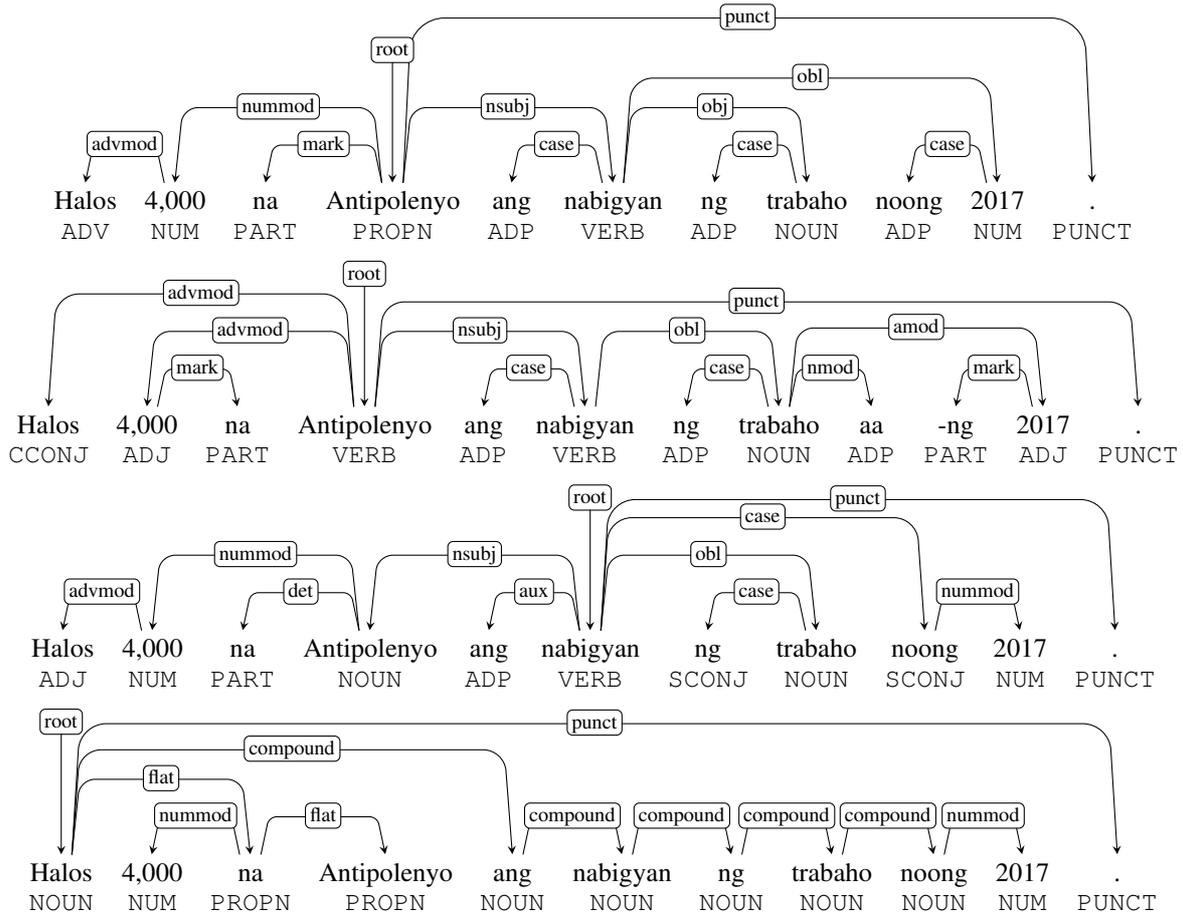

        \centering
        \begin{dependency}
            \begin{deptext}[font=\small, column sep=0.2cm]
                Halos \&
                4,000 \&
                na \&
                Antipolenyo \&
                ang \&
                nabigyan \&
                ng \&
                trabaho \&
                noong \&
                2017 \&
                .\\
                \texttt{ADV} \&
                \texttt{NUM} \&
                \texttt{PART} \&
                \texttt{PROPN} \&
                \texttt{ADP} \&
                \texttt{VERB} \&
                \texttt{ADP} \&
                \texttt{NOUN} \&
                \texttt{ADP} \&
                \texttt{NUM} \&
                \texttt{PUNCT} \\
            \end{deptext}
            \depedge{2}{1}{advmod}
            \depedge{4}{2}{nummod}
            \depedge{4}{3}{mark}
            \depedge{6}{5}{case}
            \depedge{4}{6}{nsubj}
            \depedge{8}{7}{case}
            \depedge{6}{8}{obj}
            \depedge{10}{9}{case}
            \depedge[edge unit distance=2ex]{6}{10}{obl}
            \depedge[edge unit distance=1.8ex]{4}{11}{punct}
            \deproot{4}{root}
        \end{dependency}

        \begin{dependency}
            \begin{deptext}[font=\small, column sep=0.2cm]
                Halos \&
                4,000 \&
                na \&
                Antipolenyo \&
                ang \&
                nabigyan \&
                ng \&
                trabaho \&
                aa \&
                -ng \&
                2017 \&
                .\\
                \texttt{CCONJ} \&
                \texttt{ADJ} \&
                \texttt{PART} \&
                \texttt{VERB} \&
                \texttt{ADP} \&
                \texttt{VERB} \&
                \texttt{ADP} \&
                \texttt{NOUN} \&
                \texttt{ADP} \&
                \texttt{PART} \&
                \texttt{ADJ} \&
                \texttt{PUNCT} \\
            \end{deptext}
            \depedge{4}{1}{advmod}
            \depedge{4}{2}{advmod}
            \depedge{2}{3}{mark}
            \depedge{6}{5}{case}
            \depedge{4}{6}{nsubj}
            \depedge{8}{7}{case}
            \depedge{6}{8}{obl}
            \depedge{8}{9}{nmod}
            \depedge{11}{10}{mark}
            \depedge[edge unit distance=2ex]{8}{11}{amod}
            \depedge[edge unit distance=1ex]{4}{12}{punct}
            \deproot{4}{root}
        \end{dependency}

        \begin{dependency}
            \begin{deptext}[font=\small, column sep=0.2cm]
                Halos \&
                4,000 \&
                na \&
                Antipolenyo \&
                ang \&
                nabigyan \&
                ng \&
                trabaho \&
                noong \&
                2017 \&
                .\\
                \texttt{ADJ} \&
                \texttt{NUM} \&
                \texttt{PART} \&
                \texttt{NOUN} \&
                \texttt{ADP} \&
                \texttt{VERB} \&
                \texttt{SCONJ} \&
                \texttt{NOUN} \&
                \texttt{SCONJ} \&
                \texttt{NUM} \&
                \texttt{PUNCT} \\
            \end{deptext}
            \depedge{2}{1}{advmod}
            \depedge{4}{2}{nummod}
            \depedge{4}{3}{det}
            \depedge{6}{4}{nsubj}
            \depedge{6}{5}{aux}
            \depedge{8}{7}{case}
            \depedge{6}{8}{obl}
            \depedge{6}{9}{case}
            \depedge{9}{10}{nummod}
            \depedge[edge unit distance=2ex]{6}{11}{punct}
            \deproot{6}{root}
        \end{dependency}

        \begin{dependency}
            \begin{deptext}[font=\small, column sep=0.2cm]
                Halos \&
                4,000 \&
                na \&
                Antipolenyo \&
                ang \&
                nabigyan \&
                ng \&
                trabaho \&
                noong \&
                2017 \&
                .\\
                \texttt{NOUN} \&
                \texttt{NUM} \&
                \texttt{PROPN} \&
                \texttt{PROPN} \&
                \texttt{NOUN} \&
                \texttt{NOUN} \&
                \texttt{NOUN} \&
                \texttt{NOUN} \&
                \texttt{NOUN} \&
                \texttt{NUM} \&
                \texttt{PUNCT} \\
            \end{deptext}
            \depedge{3}{2}{nummod}
            \depedge{1}{3}{flat}
            \depedge{3}{4}{flat}
            \depedge[edge unit distance=2ex]{1}{5}{compound}
            \depedge{5}{6}{compound}
            \depedge{6}{7}{compound}
            \depedge{7}{8}{compound}
            \depedge{8}{9}{compound}
            \depedge{9}{10}{nummod}
            \depedge[edge unit distance=1ex]{1}{11}{punct}
            \deproot{1}{root}
        \end{dependency}
        \caption{Ugnayan treebank annotation (1st row) and outputs of the few-shot, zero-shot, and cross-lingual pipelines (2nd to 4th rows) for the following sentence: \textit{Halos 4,000 na Antipolenyo ang nabigyan ng trabaho noong 2017} (Around 4,000 Antipolenyos were granted jobs in 2017).}
        \label{fig:pipeline-trees}
    \end{figure*}

    In addition, we find that both the few-shot and zero-shot pipelines outperform all instances of the cross-lingual pipeline by $18\%$ or more LAS. This indicates that these new pipelines improve on Tagalog parsing performance versus state-of-the-art pipelines trained on high-resource cross-lingual data (in this case, UDPipe trained on an Indonesian treebank containing over $100,000$ tokens, which yielded better LAS than several other cross-lingual alternatives); we have shown that this can be achieved through the use of alternative Tagalog language resources in both a few-shot setting (with less than $1,000$ tokens of UD-annotated training data) and a zero-shot setting (using no UD-annotated target-language data). Sample outputs for all three pipelines on news text are rendered in graphical form in Fig. \ref{fig:pipeline-trees}.

\section{Discussion}\label{sec-discussion}

    Tagalog is an Austronesian language spoken as a first language by around a quarter of the 100 million total population of the Philippines, and as a second language by the majority of the country \cite{ethnologue}. Its standardized form, Filipino, is one of two official languages for communication and instruction in the Philippines (the other being English). It has an abundance of language resources available to the public, including academic texts, creative literature, newspapers, radio \& TV broadcasts, and social media content. Furthermore, it has long been a language of interest for linguists worldwide due to its complex verbal morphology and distinctive voice system.

    Although Tagalog has both a thriving literary tradition and an established body of linguistic research behind it, computational NLP for Tagalog is only an emerging field of study, with much progress yet to be made compared to the other major languages of the world. Some of the first systems for automated grammatical analysis of Tagalog include the rule-based morphological analyzer by Fortes \shortcite{fortes-2002-constraint}, the template-based POS tagger by Rabo \shortcite{rabo-2004-template} whose tagset was the basis for the later MGNN tagset used in our tag conversion approach, and the graph-based dependency parser by Manguilimotan \& Matsumoto \shortcite{manguilimotan-matsumoto-2011-dependency}. More recent statistical and neural implementations have been tested by Go \& Nocon
    \shortcite{go-nocon-2017-using} for POS tagging, and by Yambao \& Cheng \shortcite{yambao-cheng-2020-feedforward} for morphological analysis, showing substantial improvements for each task over previous works.

    One of the shortcomings of the systems above is that their output annotations are non-standard and are not directly comparable to the results achieved for other languages. The first cross-linguistically compatible Tagalog treebank was created by Samson \shortcite{samson-2018-treebank} under the UD standard. We subsequently conducted, to our knowledge, the first comparative evaluation of supervised UD parsing for Tagalog \shortcite{aquino-de-leon-2020-parsing} using the TRG treebank by Samson as well as our own Ugnayan treebank.

    Our experiments in this paper have shown that the use of various available language resources through simple data conversion and generation methods can improve performance over supervised grammatical analysis with annotated data alone. The approaches we have tested were chosen primarily on the basis of:
    \begin{itemize}
        \item the language resources that were available to us for Tagalog;
        \item the limits in computational capacity of the consumer laptop used throughout this research; and
        \item the grammatical properties of the Tagalog language.
    \end{itemize}

    We hypothesize that the applicability of these methods to other languages will in turn be dependent on the above factors. For example, the Tagalog FST models we created for morphological analysis may be easily modified to suit other Philippine languages like Bikolano and Ilokano with similar verb infixation patterns, but would not be useful for more analytic languages like Vietnamese, and may require significant expansion for more polysynthetic languages like Kunwinjku, as seen in Lane \& Bird \shortcite{lane-bird-2019-towards}. The dependency projection method for zero-shot parsing was not effective here for Tagalog, but it may yield better results for target languages with high similarity to annotation-rich sources (e.g. Germanic and Romance languages) or if many more source languages and higher compute are used, as in Agi\'{c} et al \shortcite{agic-etal-2016-multilingual}. Language resources beyond those used in this paper, such as speech corpora and semantic networks, may require different encoding techniques and system architectures to be useful for the analysis tasks investigated here.

\section{Conclusion}\label{sec-conclusion}

    In this work, we have explored methods for leveraging alternative Tagalog language resources to improve few-shot parser performance. We have also shown that in the absence of any annotated Tagalog data, a pipeline of zero-shot methods for tokenization, tagging, and parsing also yields better results than cross-lingual models, and can even outperform the improved monolingual models when testing on out-of-domain text.

    We recommend the use of the zero-shot pipeline---consisting of unsupervised tokenization, part-of-speech tag conversion, finite-state morphological analysis, and a multilingual BERT-based parser---for general-purpose, automated grammatical analysis of Tagalog text, as it gives consistent performance for all UD annotation tasks across the two domains we have investigated here. If the text to be analyzed is sufficiently similar to the available Tagalog training data, we also recommend the use of supervised monolingual modeling, as it has been shown to yield substantial improvements over the other alternatives tested here when annotating in-domain text.

    There exist many possible avenues for building on this work, including but not limited to the creation of new treebanks and the expansion of existing treebanks, the use of domain adaptation methods for out-of-domain parsing, the exploration of Tagalog as a pivot between foreign and Philippine languages, and the investigation of methods for annotating and parsing code-switched (e.g. mixed Tagalog and English) text. We hope that our initial steps here will encourage further work towards robust grammatical analysis for Tagalog, and for many other local languages here and abroad.

\section*{Acknowledgments}

    We are grateful to Professors Mary Ann Bacolod, Anastacia Alvarez, and Rhandley Cajote for their guidance and support over the course of this project. We also thank the four anonymous reviewers for their insightful comments on an earlier draft of this paper. This work was funded by the U.P. Teaching Assistantship Program.

\bibliographystyle{acl}
\bibliography{anthology,aaa_thesis}

\end{document}